# Relative Importance of Hyperparameters in PEGASOS SVM for Imbalanced Classification by John Sun

## 1 Abstract

We aim to demonstrate in experiments that our cost sensitive PEGASOS SVM achieves good performance on imbalanced data sets with a Majority to Minority Ratio ranging from 8.6:1 to 130:1 and to ascertain whether the including intercept (bias), regularization and parameters affects performance on our selection of datasets. Although many resort to SMOTE methods, we aim for a less computationally intensive method. We evaluate the performance by examining the learning curves. These curves diagnose whether we overfit or underfit or whether the random sample of data chosen during the process was not random enough or diverse enough in dependent variable class for the algorithm to generalized to unseen examples. We will also see the background of the hyperparameters versus the test and train error in validation curves. We benchmark our PEGASOS Cost-Sensitive SVM's results of Ding's LINEAR SVM DECIDL method. He obtained an ROC-AUC of .5 in one dataset. Our work will extend the work of Ding by incorporating kernels into SVM. We will use Python rather than MATLAB as python has dictionaries for storing mixed data types during multi-parameter cross-validation.

## 2 Introduction

A class-weighted version of Shai Shalev-Shwartz et al.'s PEGASOS SVM is suited for imbalanced binary classification as it samples during stochastic subgradient descent from both predictor variable's classes with equal probability. Using minibatch of gradient descent has issues as the vector of y's in the subgradient includes both labels from class one and negative one. We did not know how to weigh a mixed-vector or define that, so we defer to SSGD (stochastic subgradient descent) where we only sample one point from the training data per optimization iteration using Python's random.choices function.

We minimize the primal objective function rather than the dual. Time complexity for an epsilon-accurate solution summed over all iterations up to where $\lambda$ is the regularization hyperparameter and $d$ refers to the dimension of the data: $\mathcal{O}(\frac{d}{\lambda\epsilon})$. We do not have to pick the step-size as we have a formula for step size from Shwartz et al. The algorithm resembles averaged-ssgd as a result of the step-size criteria. It behooves us that we do not have to pick step-size; it is a convenience in the programming and another reason to choose our method. The authors of the PEGASOS algorithm noted that omitting projection step required of subgradient descent made no negative impacts, and we follow this suggestion. We use 1000 iterations for any data. We use ROC-AUC as the criteria in cross validation. After we obtain the best hyperparameters. We report our final results on the choices that yields highest ROC-AUC.

We assume the iterations in ssgd represent independent Bernoulli trials. A success means the ROC-AUC increased.

Let $Y$, the number of independent Bernoulli trials prior to the ROC-AUC increasing relative to baseline. $X = Y + 1$ equals the number of independent Bernoulli trials before an increase in the ROC-AUC. Assume $p$ equals the probability ROC-AUC increased with respect to baseline. We estimate $p$ using maximum likelihood estimation. Our maximum likelihood estimate is $\hat{p}$. We apply probability theory to determine how many iterations to wait before checking whether the ROC-AUC increased and halting if no improvement found. With the right stopping criteria, we will train in less time. For a geometric random variable X,

$$\mu = \mathbb{E}(X) = \frac{1}{p}.$$
$$\sigma^2 = Var(X) = \frac{1-p}{p^2}$$

Assume $k = k_1,...,k_n$ be a sample where $k_i \geq 1$ for $i = 1,...,n$. $k_i$ are independent and identically distributed Geom(p). We acknowledge that the samples towards the later iterations had higher waiting times and the identically distribution assumption may not hold. For simplicity and to estimate a stopping criteria, we keep our assumption. Then $p$ can be estimated with bias:

$$\hat{p} = \left(\frac{1}{n}\sum_{i=1}^{n} k_i\right)^{-1} = \frac{n}{\sum_{i=1}^{n} k_i}.$$

We use the bias corrected estimator of p:

$$\hat{p} - \frac{\hat{p}}{1-\hat{p}}.$$

The biased corrected mle (MLEB) for the variance

$$\sigma^2_{MLEB} = \frac{1}{n-1} \frac{1-\hat{p}}{\hat{p}^2}.$$

We will grid-search within our confidence interval for the mean of the random variable X. We build a confidence interval by solving for under the idea that

$$Q_A = \frac{\sqrt{n}}{\hat{\sigma}_{MLEB}} \left(\bar{X}_n - \hat{\mu}_{MLEB}\right)$$

converges to the standard normal distribution under large sample sizes by the Central Limit Theorem. The endpoints of our confidence interval are the $\bar{X}$ s.t.

$$Q_A = \sqrt{n} \frac{\bar{X} - \hat{p} + \frac{\hat{p}}{1-\hat{p}}}{\sqrt{(1-\hat{p})/((n-1)\hat{p}^2)}} = \pm 1.96$$

The early stop parameter shapes the hyperparameters, so we will calculate the optimal stop parameter for every combination of lambda and bias prior to cross-validation. A single run of PEGASOS returns the vector of probabilities we use to calculate sample mean and variance.

## 3 Imbalanced and hyperplane separable data

Suppose $I \sim Ber(1/2)$ and $I : S \to \{-1, 1\}$.

Assume there exists elementary events $E_1, E_2, ..., E_{n-}$ such that $I(\omega) = 1, \omega \in \bigcup_{i=1}^{n_+} E_i$ and there exists elementary events $F_1, F_2, ..., F_{n_+}$ such that $I(\tau) = -1, \tau \in \bigcup_{i=1}^{n_-} F_j$.

By hypothesis:

$$P\left(\bigcup_{i=1}^{n_+} E_i\right) = \sum_{i=1}^{n_+} P(E_i) = \frac{1}{2}$$

$$P\left(\bigcup_{j=1}^{n_-} F_i\right) = \sum_{j=1}^{n_-} P(F_i) = \frac{1}{2}$$

Furthermore, we assume all elementary events have equal probability of occuring: $\forall i \leq n_-$

$$P(E_i) = \frac{1}{2n_+}.$$

$$P(F_j) = \frac{1}{2n_-}.$$

$$n_+ P(E_i) = P(I = -1) = \frac{1}{2} = n_- P(F_j) = P(I = +1).$$

Notice: $P(I = 1) + P(I = -1) = 1$.

## 3.1 Class-Weighted Objective Function

### 3.1.1 LINEAR SVM

$$\mathbf{w}, \mathbf{x} \in \mathbb{R}^d, b \in \mathbb{R}, y \in \{-1, 1\}, \lambda \in \mathbb{R}^+$$

$$\min_{w} \left( \frac{\lambda}{2} \|\mathbf{w}\|_2^2 + \sum_{i=1}^{n_+} P(E_j)[1 - (\langle \mathbf{w}, \mathbf{x}_i \rangle + b)]_+ + \sum_{i=1}^{n_-} P(F^i)[1 + (\langle \mathbf{w}, \mathbf{x}^i \rangle + b)]_+ \right)$$

### 3.1.2 Nonlinear SVM

$$\min_{\alpha} \frac{\lambda}{2} \sum_{i,j=1}^{m} \alpha[i]\alpha[j] K(\mathbf{x}_i, \mathbf{x}_j) + \sum_{i=1}^{n_+} \frac{1}{2n_+} \max\left\{0, 1 - \sum_{j=1}^{n_+} \alpha[j] K(\mathbf{x}_i, \mathbf{x}_j)\right\} + \sum_{i=1}^{n_-} \frac{1}{2n_-} \max\left\{0, 1 + \sum_{j=1}^{n_-} \alpha[j] K(\mathbf{x}^i, \mathbf{x}^j)\right\}$$

**Non-Kernel PEGASOS**

- **INPUT:** $S, T, b, \lambda, X$
- **INITIALIZE:** Set $\mathbf{w}_1 = 0$ ROC-AUC$= 0.5$
- **FOR** $t = 1, 2, \ldots, T$
  - **Pick** $\omega_t = (\mathbf{x}, y) \in S$
  - **Set** $\eta_t = \frac{1}{\lambda t}$
  - $[y(\langle \mathbf{w}, x \rangle + b) < 1]$ and $y = 1$:
    $$\mathbf{w}_{t+1} \leftarrow \mathbf{w}_t - \eta_t \nabla_t = (1 - \frac{1}{t})\mathbf{w}_t + \frac{1}{\lambda t}\frac{1}{2n_+}\mathbf{x}$$
  - $[y_i(\langle \mathbf{w}, x \rangle + b) < 1]$ and $y = -1$:
    $$\mathbf{w}_{t+1} \leftarrow \mathbf{w}_t - \eta_t \nabla_t = (1 - \frac{1}{t})\mathbf{w}_t - \frac{1}{\lambda t}\frac{1}{2n_-}\mathbf{x}$$
  - $[y_i(\langle \mathbf{w}, x \rangle + b) \geq 1]$ :
    $$\mathbf{w}_{t+1} \leftarrow \mathbf{w}_t(1 - \frac{1}{t})$$
  - If $t$ is divisible by X and ROC-AUC$(target(S), \mathbf{w}^T \textbf{features}(\mathbf{S}) + \mathbf{b}) \leq$ ROC-AUC: **Halt** and **Return weights**
  - Update ROC-AUC
  - Optional:
    $$\mathbf{w_{t+1}} \leftarrow \min\{1, \frac{1/\sqrt{\lambda}}{\|\mathbf{w_{t+1}}\|}\}\mathbf{w_{t+1}}.$$
- **OUTPUT:** $\mathbf{w}_{T+1}$

**Kernel Pegasos SVM for Imbalanced Classification Pseudocode**

- **INPUT:** $S, T, X, K(x_{it}, x_j), \lambda$
- **INITIALIZE:** Set $\alpha_1 = 0$
- **FOR** $t = 1, 2, \ldots, T$
- **Pick** $\omega_t = (\mathbf{x}, y) \in S$.
  $(\mathbf{x}, y)$ belongs to row $i_t$ of $S$.

- For all $j \neq i_t$, set $\alpha_{t+1}[j] = \alpha_t[j]$
- If $y_{i_t} = 1$ and $\frac{1}{\lambda t} \sum_{j=1}^{n_+} (\alpha_t[j] K(\mathbf{x}_{i_t}, \mathbf{x}_j)) < 1$, then: Set $\alpha_{t+1}[i_t] = \alpha_t[i_t] + \frac{1}{2n_+}$
- Else If $y_{i_t} = -1$ and $\frac{1}{\lambda t} \sum_{j=1}^{n_-} \alpha_t[j] K(\mathbf{x}^{i_t}, \mathbf{x}^j) < 1$, then: Set $\alpha_{t+1}[i_t] = \alpha_t[i_t] + \frac{1}{2n_-}$

---

- Else: Set $\alpha_{t+1}[i_t] = \alpha_t[i_t]$
- If $t$ is divisible by X and ROC-AUC$(target(S), \sum_{j=1}^{m} \alpha_{t+1}[j] y_j K(x_i, x_j) + b) \leq$ ROC-AUC: **Halt** and **Return weights**
- Update AUC-ROC
- OUTPUT: $\alpha_{T+1}$

## 4 Data

We use the collection of datasets requiring no pre-processing from Ding (2011).

Table 1: Data

| Dataset Name | Target | Rows in Data | Minority/Majority | Total Features |
|---|---|---|---|---|
| Ecoli | imU | 336 | 8.6:1 | 7 |
| Oil | minority | 937 | 22:1 | 49 |
| $Abalone_{19}$ | 19 | 4177 | 130:1 | 10 |
| ozone level | ozone | 2536 | 34:1 | 72 |
| spectrometer | $\geq 44$ | 531 | 11:1 | 93 |
| mammography | minority | 11183 | 42:1 | 6 |
| Yeast $ML_8$ | 8 | 2417 | 13:1 | 103 |

## 5 5-Fold Cross-Validation Linear Kernel on 80 % Partition

Table 2: Best Parameters for Linear PEGASOS SVM

| Dataset ID | Mean Areas under ROCs | Regularization $\lambda$ | Bias | Optimal Trials to AUC-ROC improvement |
|---|---|---|---|---|
| Ecoli | .9283 | .0001 | 1.1111 | 5 |
| Oil | .70 | .1 | 1.55 | 6 |
| $Abalone_{19}$ | .71 | .01 | -1.111 | 4 |
| ozone level | .624 | 1 | .222 | 2 |
| spectrometer | .668 | 1.0 | -2/3 | 3 |
| mammography | .9269 | 1 | -1.111 | 8 |

# 5 Diagnosis of Cross-Validation

## 5.1 Linear SVM

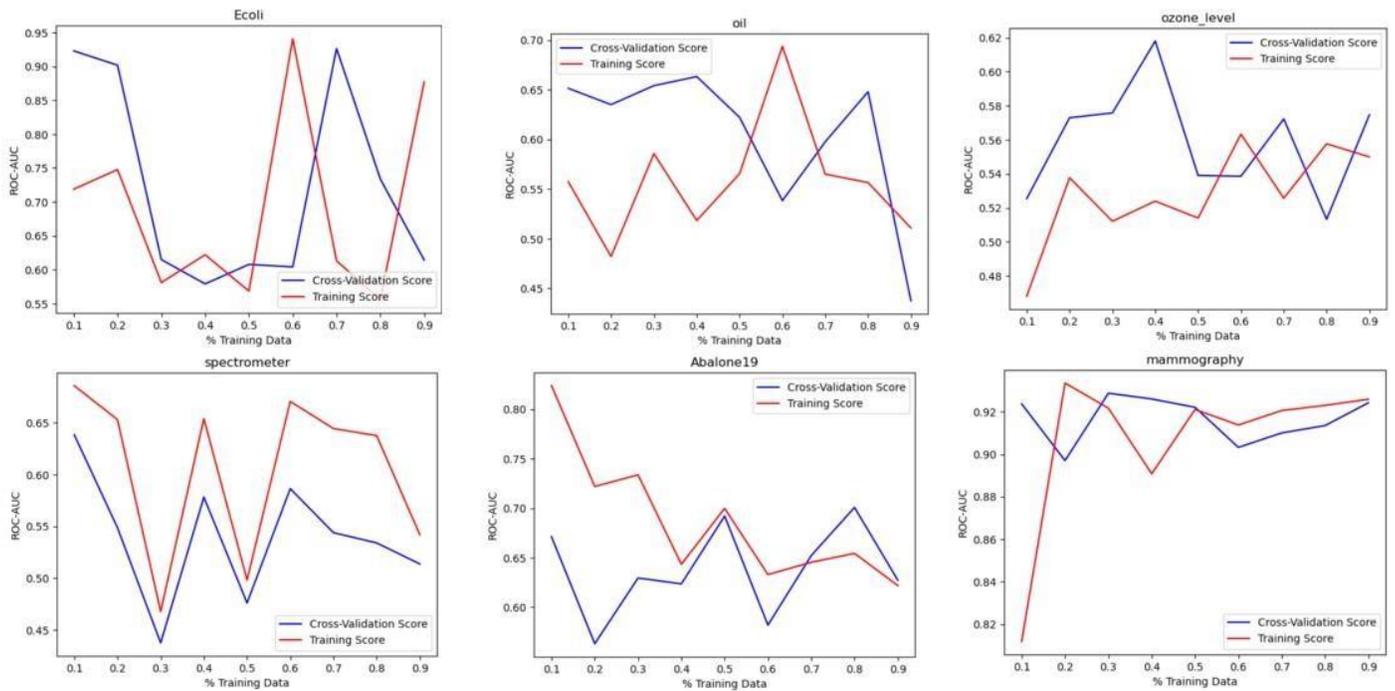

We choose 5-fold as we hypothesized splits of training on 80 percent would be reasonable for practical speed uses. To generate these curves we ran 5-fold cross-validation on different proportions of training data. To summarize, the curves when divergent from one another indicate either overfitting or underfitting. We noticed a strong dependency on our early stopping criteria affects the performance.

We have a good learning curve on the Ecoli and mammography dataset. In the good cases, we had both low bias and variance as we had nearly perfect AUC and the cross-validation score curves and training curves coincided.

However, not all was good on two extremes. For example, based on the curves, the PEGASOS SSGD algorithm overfitted the training examples randomly selected from the Ozone level dataset. The parameters estimated from the training sample on the spectrometer dataset underfit the out-of-training data as the training score curve oscillated rather than decreased as a result of varying the independent variable % training data. We aim to study or find the data conditions or algorithms that will overfit, the Spectrometer dataset, and when there's a decrease, increase or invariance to change in that independent variable in future studies.

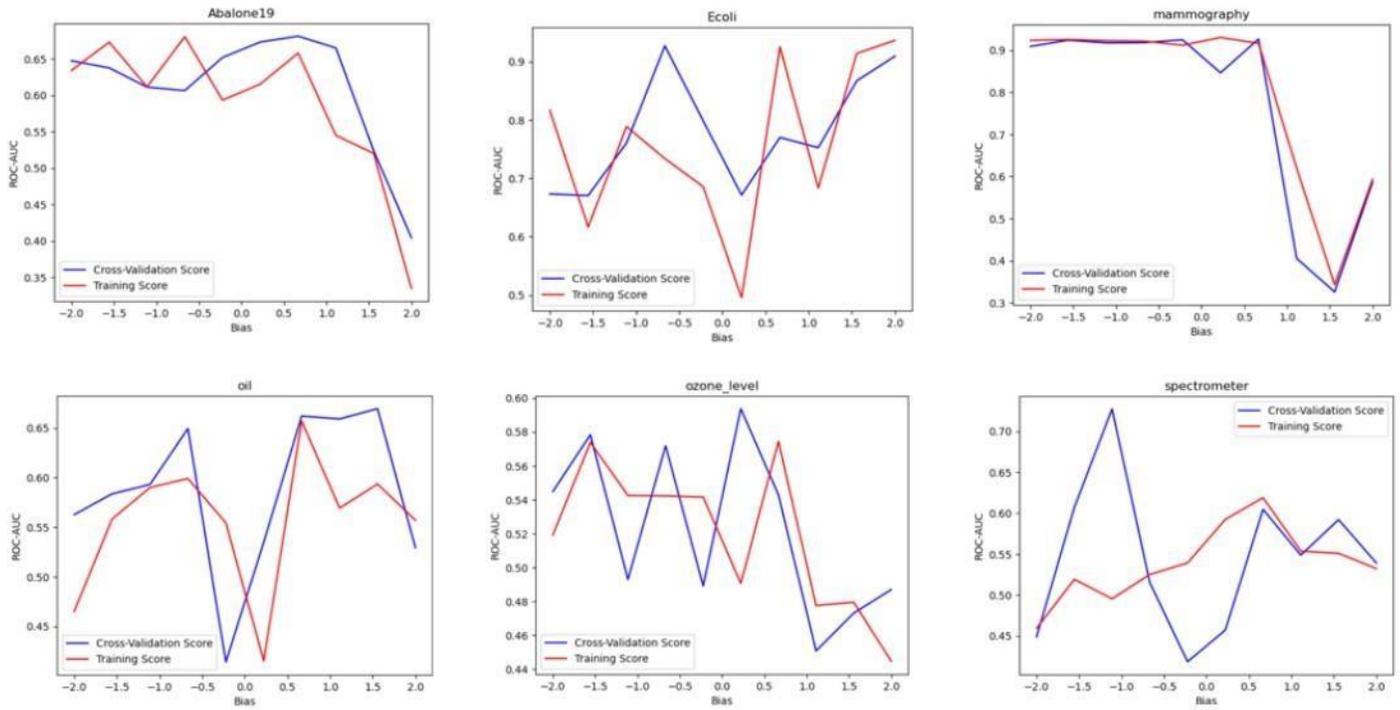

We ran five-fold validation for various biases (inclusion of an intercept in the predictive model) via a grid-search. As expected, bias hyperparameter determined performance in most cases. Higher biases led to worse performance except on the Ecoli dataset. Most datasets depended on this bias term in the prediction equation. However, in the Mammography dataset, the inclusion of a bias term had no effect on the performance when the bias (intercept) ranged from negative two to one-half. As a lesson learned, we predict on future prediction on dataset similar to those used in this study, one should drop the bias term as it leads to a computationally simpler, faster, and more memory efficient solution.

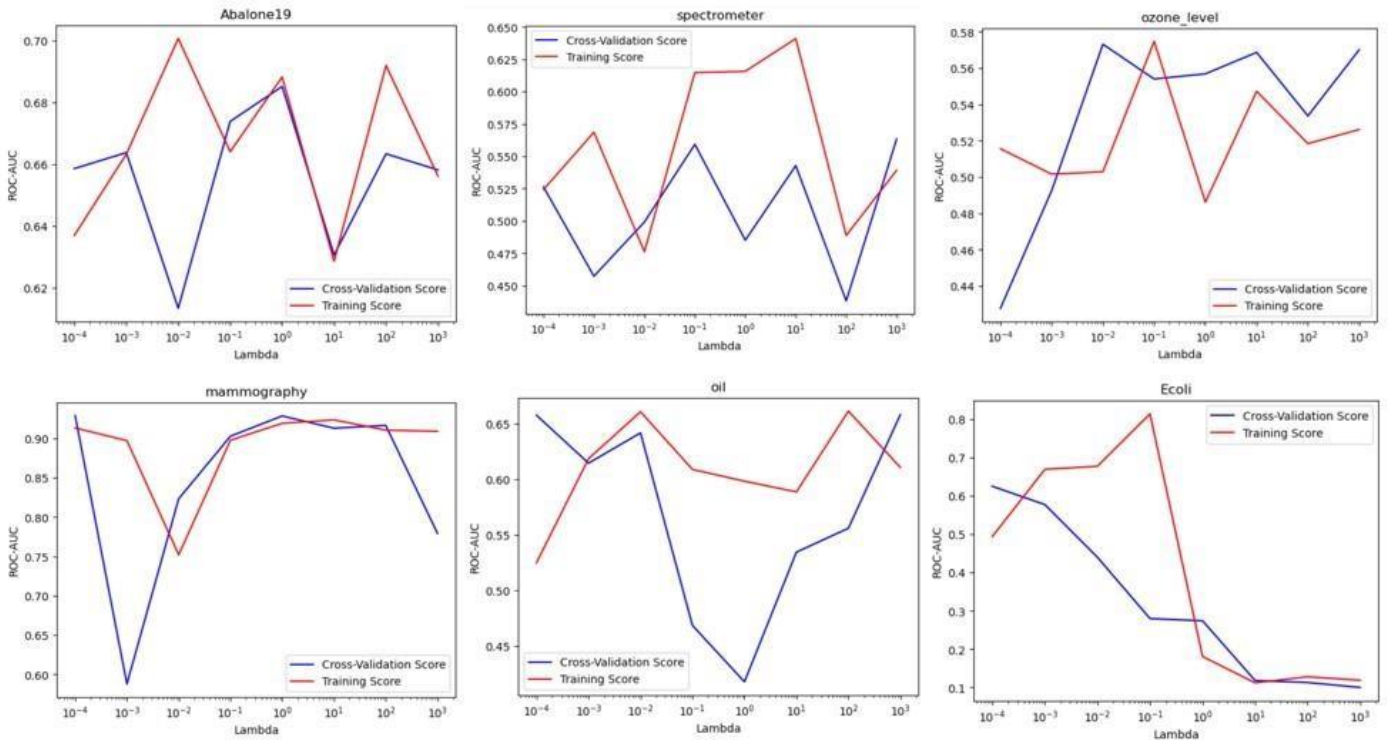

We believe this graph confirms the original design to use include a regularization term. The regularization term supported good performance on Ecoli dataset, the spectrometer dataset, the abalone19 dataset. Future studies on data similar to ours should include the regularization term or increase the search space of the regularization parameter. Increasing the search space of the regularization parameter with the knowledge that the bias term is unnecessary may lead to prediction accuracy and ROC-AUC score gains.

# 6 Metrics we Considered for Imbalanced Classification

Halimu et al. (2019) reported that ROC's AUC found the more discriminating model in a experiment on imbalanced and balanced datasets among several tests in inclusion of Matthew's correlation coefficient (MCC) versus the area under the curve of the receiver operator characteristic curve using SVM and other algorithms. We chose to not use Matthew's correlation coefficient as it led to undefined scores on several of the datasets, we had planned to study prior to conducting of the actual experiments.

The area under curve of the receiver operator characteristic curve in discrete case is

$$\frac{1}{n_+ n_-} \sum_{i,j=1} (f(x_i^+) > f(x_j^-))$$

.

$f(x_i^+)$ equals the prediction score of $x_i$ whose label belongs to positive class. $f(x_j^-)$ equals the prediction score of $x_j$ that belons to the negative class. We relied of Sci-Kit learn's AUC function which plots the ROC curve and uses a trapezoidal approximation to find the AUC.

We noticed that we required a high-roc to have accurate predictions in both classes. With auc lower than .8, we noticed either all correct in minority class and mostly wrong in the majority class- a failure to perform good imbalanced classification. This may suggest ROC-AUC is a bad metric as scores traditionally accepted in the medical field of (.8) are unreliable for these datasets but in a Machine Learning context.

We perform cross-validation in linear PEGASOS SVM with hyperparameters $\lambda \in \{0.0001, .001, .01, .1, 1, 10, 100, 1000\}$. and hyperparameter bias terms in a linear space of 10 elements starting at negative two, ending at positive two.

# 7 DECIDL Results QP

## 8.1 Linear Kernel

Table 3: SVM

| Dataset Name | ROC-AUC |
|---|---|
| Ecoli | .924 |
| Oil | .884 |
| $abalone_{19}$ | .743 |
| ozone level | .865 |
| spectrometer | .97 |
| mammography | .854 |
| Yeast $ML_*8$ | .5 |

# 9 Our Results on 20 % Holdout

Table 4: LINEAR SGD Cost-Sensitive PEGASOS

| Dataset Name | ROC-AUC |
|---|---|
| Ecoli | .829 |
| Oil | .742 |
| $abalone_{19}$ | .666 |
| ozone level | .407 |
| spectrometer | .459 |
| mammography | .861 |

# 8 Weights of Kernel Pegasos

The dual variable stands for the LaGrange multiplier in machine-learning in optimization. In fact the alphavariable is often a real number. Yet, unusually in Kernel PEGASOS, the vector of natural numbers of alphas counts the number of times the weights increased the empirical risk of the loss function in training. The tuple of $\alpha$ reaches stability as the weights will eventually incur zero loss in the limit. This update rules insists that we use some piecewise loss function that has a zero part and non-zero part.

# 9 Other methodology: Over and Under-Sampling Methods for Treating Imbalanced Data

SMOTE (Synthetic majority oversampling technique) introduces additional parameters to cross-validate than the bias, kernel hyperparameter, regularization parameter for SVM. This complexity makes computation more difficult. Some other algorithms introduce a number of neighbors parameter. In this, a random generator samples an equal number of items from the dependent variable's classes, so we already a type of oversampling. While the SMOTE method uses several hyperparameters, our imbalanced PEGASOS SVM should in theory have less computational complexity than the SMOTE method as we cross-validate fewer parameters.

## 9.1 History of SMOTE

In the early developments one trains on an oversampled by random selection or complete selection of the minority class from the original data. The random selection creates synthetic data not representative of the original as a result of subsetting and leads to overfitting, so one would have better performance using a complete sample of the minority classed data.

Many other methods create duplicates of data to address the lack of a addressing of the issue in the machine learning algorithm. For example, borderline Smote, MWMOTE, and ADSYN sample the minority classed data in a complete fashion along the decision boundaries between the two classes. Unlike its predecessors, the SMOTE method does not duplicate existing data, but rather takes one point and its nearest neighbor from the minority class, and generates a new datapoint along a convex combination of the two (Ye et al, 2020). Experimentation found that generating from sampling points of the minority classed data from the decision boundary rather than away from the decision balanced helped train the model training.

Kernel SMOTE and LLE Smote developed after ADSYN and MWMOTE later identified an issue resulting from generating new points lying on a convex combination when the majority and minority classes had nonlinearly separable dependent variable decision boundary. The generated point represented noise rather than signal. Both Kernel SMOTE and LLE Smote map the data into a new space, and have adjustable hyperparameters tunable for creating the mapped data separable by a hyperplane. LLE SMOTE suits the majority of cases where one would forgo the Kernel Support Vector Machine method as it does not allow for SVM kernels. For a blackbox implementation of these SMOTE methods see *smote-variants: a Python Implementation of 85 Minority Oversampling Techniques*. Among these 85, polynom-fit-SMOTE yields the highest AUC average of .90258 in a competition involving around 100 datasets.

# 12 Appendix

## 12.1 Example

$$m = 10, n_+ = 7, n_- = 3$$

$$P(E_i) = \frac{1}{6}$$

$$P(F_j) = \frac{1}{14}$$

$$P(I = 1) + P(I = -1) = 1$$

$$\min_{\mathbf{w}} \left( \frac{\lambda}{2} ||\mathbf{w}||_2^2 + \sum_{(\mathbf{x},y) \in S^+} \frac{1}{14}[1 - (\langle \mathbf{w}, \mathbf{x} \rangle + b)]_+ + \sum_{(\mathbf{x},y) \in S^-} \frac{1}{6}[1 + (\langle \mathbf{w}, \mathbf{x} \rangle + b)]_+ \right)$$

## 12.2 Weight Update Derivation in Pegasos SVM

### 12.2.1 $[y_i(\langle \mathbf{w}, x \rangle) + b) \geq 1]$

$$w_{t+1} = w_t(1 - \frac{1}{t})$$

### 12.2.2 $[y_i(\langle \mathbf{w}, x \rangle) + b) < 1]$ and $y = 1$

$$w_{t+1} = w_t - \eta_t \nabla_t$$

$$w_{t+1} = w_t - \frac{1}{\lambda t}(\lambda w_t - P(F_j) y_{i_t} x_{i_t})$$

$$w_{t+1} = w_t - \frac{w_t}{t} + \frac{1}{\lambda t} P(F_j) y_{i_t} x_{i_t}$$

$$w_{t+1} = w_t(1 - \frac{1}{t}) + \frac{1}{\lambda t} P(F_j) y_{i_t} x_{i_t}$$

### 12.2.3 $[y_i(\langle \mathbf{w}, x \rangle) + b) < 1]$ and $y = -1$

$$w_{t+1} = w_t - \eta_t \nabla_t$$

$$w_{t+1} = w_t - \frac{1}{\lambda t}(\lambda w_t - P(E_i) y_{i_t} x_{i_t})$$

$$w_{t+1} = w_t - \frac{w_t}{t} + \frac{1}{\lambda t} P(E_i) y_{i_t} x_{i_t}$$

$$w_{t+1} = w_t(1 - \frac{1}{t}) + \frac{1}{\lambda t} P(E_i) y_{i_t} x_{i_t}$$